\documentclass[final,5p,times,twocolumn,fleqn]{elsarticle}
\journal{Optics and Lasers in Engineering}

\usepackage{url}
\usepackage{graphicx}
\usepackage{subfigure}
\usepackage{amsmath,bm}
\usepackage{amssymb}
\usepackage{multirow}
\usepackage{caption}
\usepackage[table]{xcolor}
\usepackage{diagbox}
\usepackage{booktabs}
\usepackage{algorithm}  
\usepackage{algorithmicx}  
\usepackage{tabularx}
\usepackage{threeparttable}
\makeatletter
\renewcommand{\textbf}{\bfseries}
\title{\LARGE \bf Learning Inter- and Intraframe Representations for Non-Lambertian Photometric Stereo}

\begin{document}

\begin{frontmatter}

\author[a,b]{Yanlong Cao}
\author[a,b]{Binjie Ding} 
\author[a,b]{Zewei He} 
\author[a,b]{Jiangxin Yang}
\author[a,b]{Jingxi Chen}
\author[a,b]{Yanpeng Cao\corref{cor1}}
\cortext[cor1]{Corresponding author}
\ead{caoyp@zju.edu.cn}

\author[c]{Xin Li}

\address[a]{State Key Laboratory of Fluid Power and Mechatronic Systems, School of  Mechanical Engineering, Zhejiang University, Hangzhou, China}
\address[b]{Key Laboratory of Advanced Manufacturing Technology of Zhejiang Province, School of Mechanical Engineering,  Zhejiang University, Hangzhou, China}
\address[c]{School of the Electrical Engineering and Computer Science (EECS), Louisiana State University, Baton Rouge, LA 70803, USA} 


\begin{abstract}
Photometric stereo provides an important method for high-fidelity 3D reconstruction based on multiple intensity images captured under different illumination directions. In this paper, we present a complete framework, including a multilight source illumination and acquisition hardware system and a two-stage convolutional neural network (CNN) architecture, to construct inter- and intraframe representations for accurate normal estimation of non-Lambertian objects. We experimentally investigate numerous network design alternatives for identifying the optimal scheme to deploy inter- and intraframe feature extraction modules for the photometric stereo problem. Moreover, we propose utilizing the easily obtained object mask to eliminate adverse interference from invalid background regions in intraframe spatial convolutions, thus effectively improving the accuracy of normal estimation for surfaces made of dark materials or with cast shadows. Experimental results demonstrate that the proposed masked two-stage photometric stereo CNN model (MT-PS-CNN) performs favourably against state-of-the-art photometric stereo techniques in terms of both accuracy and efficiency. In addition, the proposed method is capable of predicting accurate and rich surface normal details for non-Lambertian objects of complex geometry and performs stably given inputs captured in both sparse and dense lighting distributions.

\end{abstract}


\end{frontmatter}

\section{Introduction}

In recent years, photometric stereo has received significant attention in the field of optical engineering and advanced manufacturing \cite{song2018photometric,villa2004surface}. Photometric stereo techniques estimate accurate and highly detailed surface normals of a target object based on a set of images captured in different light directions using a viewpoint fixed camera. These techniques can generate 3D models with rich details to facilitate various applications, such as automated industrial quality inspection \cite{3dDetection}, high-fidelity 3D reconstruction/modelling \cite{ma2019calibration,xie2015practical}, and face recognition and verification \cite{zhou2019digital,zhou20193}.

The basic theory of photometric stereo was first proposed by Woodham et al. based on the assumption of ideal Lambertian reflectance \cite{Woodham1980}. However, most of the real-world objects are non-Lambertian. Therefore, many researchers have attempted to utilize flexible surface reflection functions and bidirectional reflection distribution functions (BRDFs) to develop more applicable photometric stereo techniques that work well for real-world objects \cite{Goldman2005,Shi2014}.

\begin{figure*}\centering
	\includegraphics[width=17cm]{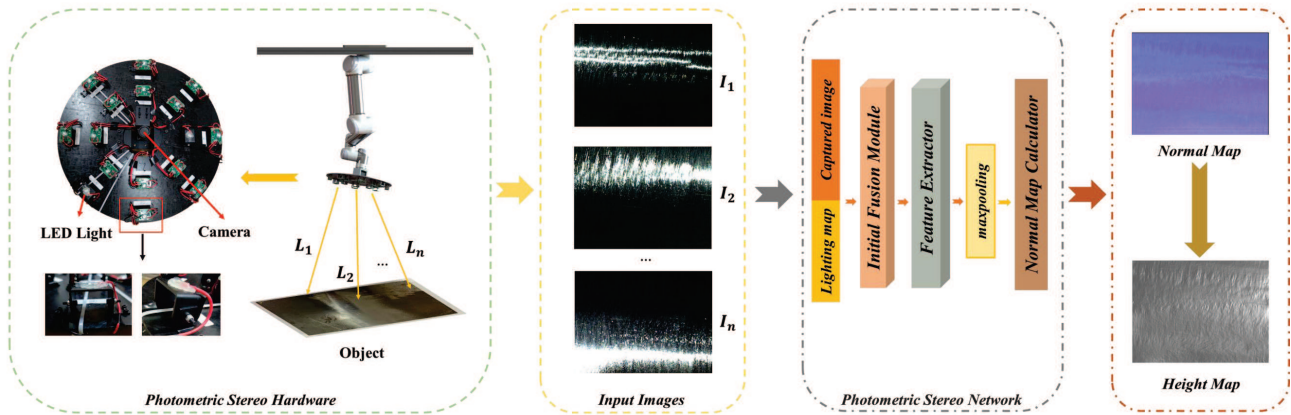}
	\caption{The data acquisition and processing workflow of our proposed photometric stereo system.}
	\label{fig:summarize}
\end{figure*}

In recent years, numerous deep learning-based methods have been proposed for high-quality photometric stereo tasks. Some approaches have attempted to explore the per-pixel intensity variation among different images (interframe clues \cite{Ikehata2018ECCV-CNN-PS}) to generate high-accuracy surface normal estimation results. However, these interframe photometric stereo methods neglect local spatial intensity variation among neighbouring pixels, encode important cues for predicting surface normals, and thus perform unsatisfactorily when the number of input images decreases. To overcome the limits of interframe photometric stereo techniques, some researchers took advantage of per-frame intensity variation among neighbouring pixels (intraframe clues \cite{Chen2018ECCV_PS-FCN,Chen2019CVPR}) to robustly estimate the surface normals when a limited number of input images were available. It is worth noting that the intraframe photometric stereo technique typically performs unfavourably compared with those methods built on interframe analysis. Therefore, it is highly desirable to develop a unified method that performs both inter- and intraframe analyses for the challenging surface normal estimation task.

In this paper, we present a complete photometric stereo data acquisition and processing framework, as shown in Fig.~\ref{fig:summarize}, constructing inter- and intraframe feature representations based on an arbitrary number of unordered images captured under different lighting configurations for high-quality surface normal estimation of non-Lambertian objects. The image acquisition system is installed on the end of a robot manipulator for image capture of target objects of various sizes. Sixteen LED lights are evenly distributed around a viewpoint-fixed camera to illuminate the target object from different directions for image capture. Given an arbitrary number of images captured under different lighting configurations, we further present a novel two-stage CNN architecture to explore valuable information encoded in both local image patches and cross adjacent frames and construct inter- and intraframe feature representations for high-quality surface normal estimation of non-Lambertian objects, as illustrated in Fig.~\ref{fig:network}. More specifically, we utilize $M\times1\times1$ convolutional layers to analyse lighting variations of individual pixels on $M$ frames and $1\times N\times N$ convolutional layers to capture intraframe intensity variation among $N\times N$ local pixels. We experimentally evaluate the performances of various network design alternatives in an attempt to identify the optimal sequence and strategy to deploy interframe and intraframe feature extraction modules.
Two important findings are noted: (1) it is better to first perform interframe feature extraction followed by intraframe feature extraction since the frame-to-frame observations provide important information for the photometric stereo task; (2) it is desirable to divide the entire feature extraction process into two individual stages, whereas mixing the inter- and intraframe feature extraction steps will adversely affect the performance of surface normal estimation. Moreover, we propose utilizing the easily obtained object mask to eliminate adverse interference from invalid background regions in intraframe spatial convolutions, thus effectively improving the accuracy of normal estimation for surfaces with insufficient reflectance observations (e.g., made of highly absorptive dark materials or with cast shadows). Compared with the state-of-the-art CNN-based photometric stereo techniques \cite{Chen2018ECCV_PS-FCN,Ikehata2018ECCV-CNN-PS}, our proposed MT-PS-CNN is capable of estimating more accurate surface normals using fewer parameters and can perform consistently well across various image capturing configurations. This work provides the following three main contributions.

\begin{itemize}
	\item  We design a two-stage CNN architecture to construct inter- and intraframe representations for photometric stereo and establish ablation studies to identify the optimal scheme to deploy interframe and intraframe feature extraction modules to achieve high-quality surface normal estimation.
	
	\item We propose utilizing the easily obtained object mask to eliminate adverse interference from invalid background regions during intraframe spatial convolutions, which provides an effective technique to facilitate accurate normal estimation for surfaces made of highly absorptive dark materials or with cast shadows. 
	
	\item With fewer parameters, our proposed MT-PS-CNN outperforms state-of-the-art photometric stereo techniques \cite{Chen2018ECCV_PS-FCN,Ikehata2018ECCV-CNN-PS}. Moreover, the proposed method is capable of predicting accurate and rich surface normal details for non-Lambertian objects and performs well with sparse input frames.
	
\end{itemize}

\section{Related Work}
\label{Related Work}
In this section, we provide a brief overview of conventional and deep-learning based photometric stereo methods for non-Lambertian objects. For a detailed introduction of recent studies of photometric stereo, readers can refer to \cite{Ackermann2015}.

\subsection{Conventional methods}

The original photometric stereo method \cite{Woodham1980} works based on the ideal Lambertian reflectance model and analyses per-pixel lighting observation variations. However, most real-world objects cannot satisfy the ideal Lambertian reflectance model; thus, non-Lambertian photometric stereo methods have been extensively researched given their increased practicability. In general, existing non-Lambertian methods can be classified into three major categories.

The first category includes robust-estimation-based approaches, which treat the non-Lambertian reflectances as outliers. 
The methods assume that the majority of observations conform to the Lambertian model and that the non-Lambertian reflectances are sparse and local \cite{Solomon1992,Coleman1982}. Recently, studies have explored relevant approaches, e.g., rank-minimization-based  methods \cite{Wu2011}, using RANSAC schemes \cite{mukaigawa2007analysis} and taking median-based approaches \cite{Miyazaki2010}. Due to the limitation of the  assumption, robust estimation-based methods require numerous inputs and experience difficulties with the thickset non-Lambertian surfaces, i.e., board shadowed areas or extensive specularity.

The second category includes reflectance model-based approaches, which arrange parametric or nonparametric models to precisely represent the appearance of real-world materials. These sophisticated reflectance models can be divided into physically based (e.g., Torrance-Sparrow model \cite{Torrance1967,Kay1995,Georghiades2003}, Cook-Torrance model \cite{Cook1981} and the Ward model \cite{Hin-ShunChung2008,Goldman2010}) and empirical models (e.g., Phong model \cite{Phong1975}, Blinn-Phong model \cite{Blinn1977}). However, these sophisticated reflectance model-based methods are only applicable for targets made of limited classes of materials \cite{ngan2005experimental,stark2005barycentric}.

The third category includes example-based methods, recovering the surface normal with an additional reference object.
Hertzmann and Seitz proposed an example-based method \cite{hertzmann2005example} that uses orientation consistency to reconstruct the surface normals of the target object.
Orientation consistency means that two points with the same surface orientation should have the same or similar appearance in an image. The example-based method can obtain the surface without solving a complex optimization; however, it requires a known-surface-parameter reference object that is typically not available during practical image acquisition.

\subsection{Deep-learning-based methods}
Due to the adovementioned limitations of conventional photometric stereo methods, deep learning techniques with the ability to approximate highly nonlinear mappings have been recently utilized to solve complex photometric stereo problems. Santo et al. \cite{Santo2017} first solved the photometric stereo task via deep learning and proposed a pixelwise method that utilizes a fully connected network (DPSN) to establish a mapping from given observations to the surface normal. The assumption that light directions must be predefined and remain the same during the training and test phases restricts the application of DPSN. Ikehata et al. \cite{Ikehata2018ECCV-CNN-PS} introduced a CNN-based method, CNN-PS, which relaxes the limitation of lighting information and image structure through a novel network input (2-D observation map). A synthetic photometric stereo dataset, called Cycles PS, was presented in this paper, considering the global illumination effects. CNN-PS performs significantly better on the benchmark dataset than most traditional photometric stereo approaches. Chen et al. \cite{Chen2018ECCV_PS-FCN} first proposed a fully convolutional network called PS-FCN that also takes unstructured images and lighting information as input. PS-FCN takes the full-size images into account and effectively utilizes the spatial information among local pixels, and this process is neglected in the previously proposed pixelwise method. Thus, the frame-based PS-FCN method performs better with a sparse input setting than the pixelwise methods. Then, Chen et al. \cite{Chen2019CVPR} extended PS-FCN and proposed an uncalibrated PS method (SDPS-Net) for non-Lambertian surfaces. SDPS-Net contains two processing steps: a classification network (LCNet) is used to approximate the light information, and a subsequent prediction network (NENet) is utilized to estimate the surface normal.

Given sufficient input images, pixelwise methods (e.g., CNN-PS \cite{Ikehata2018ECCV-CNN-PS} and DPSN \cite{Santo2017}) are generally more accurate for surface normal estimations than framewise methods (e.g., PS-FCN \cite{Chen2018ECCV_PS-FCN} and SDPS-Net \cite{Chen2019CVPR}). However, a sparse input setting will adversely decrease the performance of pixelwise methods due given that these methods ignore spatial information. To handle sparse input, Li et al. \cite{Li2019} proposed a deep learning pixelwise approach, which applies a connection table that can select relatively effective light directions, to minify the input of pixelwise photometric stereo methods. Zheng et al. \cite{Zheng2019_SPLINE-Net} proposed SPLINE-Net, which generates dense lighting observations by lighting interpolation, to improve the performance of sparse photometric stereo. Wang et al. \cite{FuLin2020} proposed a photometric stereo network that utilizes collocated light images as supplementary information to improve the performance.

\section{Approach}
\label{Approach}

Given $q$ RGB images of a target object with $p = W \times H$ pixels ($W$ and $H$ are the width and height of input images, respectively) captured under different light directions, surface normals of $p$ pixels $\boldsymbol N \in \mathbb{R}^{3\times p}$, and light directions of $q$ images $\boldsymbol L \in \mathbb{R}^{3\times q}$, the observation matrix $\boldsymbol I \in \mathbb{R}^{3\times p \times q}$ can be formulated as follows\cite{Chen2018ECCV_PS-FCN}:
\begin{equation}
	\boldsymbol I = \boldsymbol\Theta \circ \text{repmat}(\max(\boldsymbol N^{\top} \boldsymbol L,0),3),
	\label{Eqn2}
\end{equation}
where $\boldsymbol\Theta \in \mathbb{R}^{3\times p\times q}$ is a complex function of the surface normal, light direction, and viewing direction (the viewing direction is set to $[0, 0, 1]^\top$, which is parallel to the z-axis of the world coordinates), and $\circ$ represents the elementwise dot product. In this paper, we present a two-stage CNN model MT-PS-CNN to extract inter- and intraframe feature representations to directly estimate the normal matrix $\boldsymbol N$ based on the observation matrix $\boldsymbol I$ and light direction matrix $\boldsymbol L$ without explicitly modelling the complex $\boldsymbol \Theta$ function.


\begin{figure}[t]\centering
	\includegraphics[width=8cm]{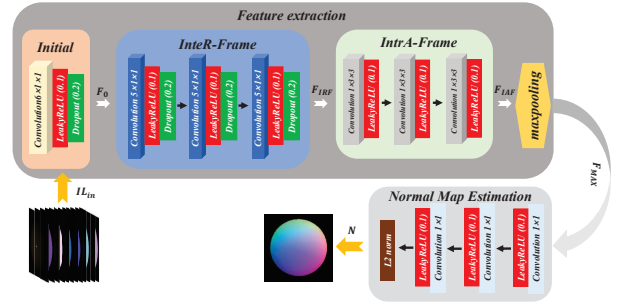}
	\caption{The overall architecture of the proposed MT-PS-CNN model.}\label{fig:network}
\end{figure}

\subsection{Network architecture}

As illustrated in Fig.~\ref{fig:network}, our proposed MT-PS-CNN model consists of three major components: initial feature extraction, inter- and intraframe feature extraction, and normal map estimation. For each captured image, we follow conventional practice to replicate its corresponding 3-vector light direction along the spatial directions to obtain a $3\times H \times W$ lighting map \cite{Chen2018ECCV_PS-FCN}. We apply the binary object mask to the lighting map, extract light directions for the target object (mask pixel values equal to 1) and eliminate invalid background regions (mask pixel values equal to 0). More details on generating 2D masked lighting maps are provided in Sec.~\ref{masked_light_map}. The obtained 3-channel masked lighting map is concatenated with the 3-channel RGB image to generate a 6-channel image-light data matrix. We stack $q$ image-light matrices together to obtain the input of our MT-PS-CNN $\boldsymbol{IL}_{input}$ which is a $(6 \times q)\times H\times W$ data matrix. Note that we purposely store input images and masked lighting masks in a 2D matrix for the following spatial convolutions.

Given input $\boldsymbol{IL}_{input}$, we first deploy the initial feature extraction module to compute feature map $F_{0}$ as follows:
\begin{equation}
	F_{0} = \mathcal{F}_{dropout}(\sigma(Con_{6\times 1\times 1}(\boldsymbol{IL}_{input})),
\end{equation}
where $Con_{6\times 1\times 1}(\cdot)$ denotes a 3D convolutional layer of the $6\times1\times1$ kernel. Note that the kernel size of 3D convolution is set to $6\times1\times1$ to process the concatenated 6-channel image-light data matrices. Following the 3D convolutional layer, a leaky ReLU layer $\sigma(\cdot) = max(0.1x,x)$ and a dropout layer $\mathcal{F}_{dropout}(\cdot)$ are deployed to activate the values and simulate the cast shadow effects \cite{Santo2017,Ikehata2018ECCV-CNN-PS}.

Then, we deploy a number of inter- and intraframe feature extraction blocks to perform simultaneous analysis of frame-to-frame and per-frame lighting variations. Within each InteR-frame Feature Extraction (IRFE) block, we utilize 3D convolutional layers of kernel size $M\times1\times1$ (i.e., $Con_{M\times 1\times 1}(\cdot)$) to process lighting variations of individual pixels on $M$ adjacent frames. Such interframe information provides important clues to eliminate the influence of outliers (i.e., shadows, interreflection, and specularity, etc.) and thus leads to accurate restoration results in photometric stereo tasks \cite{Mukaigawa2007}. Each convolution is followed by a leaky ReLU activation and a dropout layer. We stack $K$ IRFE blocks to compute interframe feature representations $F^{K}_{IRF}$ as follows:
\begin{align}\centering
	F^{1}_{IRF} = &\; IRFE^{1}\left(F_0\right)\\
	F^{2}_{IRF} = &\; IRFE^{2}\left(F^{1}_{IRF}\right)\\
	\nonumber &\vdots\\
	F^{K}_{IRF} = &\; IRFE^{K}\left(F^{N-1}_{IRF}\right),
\end{align}
where ${IRFE}^{i}(\cdot)=\mathcal{F}_{dropout}(\sigma(Con_{M\times1\times1}^{i}(\cdot))$ represents the operations of the $i_{th}$ IRFE block. 

The computed feature map $F^{K}_{IRF}$ is then fed to a number of IntrA-frame feature extraction (IAFE) blocks to exploit the spatial information in local image patches and compute interframe feature representations $F^{L}_{IAF}$ as follows:

\begin{align}
	F^{1}_{IAF} = &\; IAFE^{1}\left(F^{K}_{IRF}\right)\\
	F^{2}_{IAF} = &\; IAFE^{2}\left(F^{1}_{IRF}\right)\\
	\nonumber&\vdots\\
	F^{L}_{IAF} = &\; IAFE^{L}\left(F^{L-1}_{IRF}\right),
\end{align}
where $L$ is the total number of stacked IAFE blocks, and ${IAFE}^{i}(\cdot)=\sigma(Con_{1 \times N \times N}^{i}(\cdot))$ denotes the operations of the $i_{th}$ IAFE block. Note that each IAFE block contains 3D convolutional layers of kernel size $1 \times N \times N$ (i.e., $Con_{1 \times N \times N}(\cdot)$) and leaky ReLU activation to capture intraframe intensity variation among $N\times N$ local pixels. The extracted local context information can improve the performance of CNN models to handle various reflectances and performs robustly under sparse lighting distributions \cite{Li2019CVPR,Wang2020TIP}. In our implementation, we experimentally set $K=L=3$ to achieve a good balance between model complexity and good performance.

To handle a flexible number of input images in photometric stereo tasks, order-agnostic operations (e.g., pooling layers) \cite{Hartmann2017learned,Wiles2017silnet} are typically utilized to standardize/fix the channel number of feature maps. Following the research work of Chen et al. \cite{Chen2018ECCV_PS-FCN}, we apply the max-pooling operation ($MP$) to compress the channel number of $F^{L}_{IAF}$ as follows: 
\begin{equation}
	F_{MAX} = MP (F^{L}_{IAF}),
\end{equation}
where $F_{MAX}$ is the output of the max-pooling operation. It is a representation with a fixed number of channels by aggregating most salient features from images captured under different light directions. 

A normal map estimation subnetwork is appended after the max-pooling operation for normal map estimation, converting the computed feature $F_{MAX}$ to the surface normal $N_{i,j}$ ($i$ and $j$ denote the spatial coordinates of the normal map) as follows:
\begin{align}
	&N_{i,j} = NME (F_{MAX}), \nonumber\\
	&= L^2_{norm} (\sigma(Con_{1\times1} (\sigma(Con_{1\times1} (\sigma(Con_{1\times1}(F_{MAX}))))))),	
\end{align}
where $NEM(\cdot)$ denotes the operations of the Normal Map Estimation module, which contains three $1\times 1$ convolutional layers, two leaky ReLU activations and an L2-normalization layer $L^2_{norm}$ for predicting the normal map.

The training of the proposed MT-PS-CNN model is driven by minimizing the error between the predicted and ground truth normal maps. We adopt the commonly used cosine similarity loss, which is formulated as follows:
\begin{equation}
	\mathcal{L}=\frac{1}{HW}\sum_{i,j}(1-N_{i,j} \cdot \tilde{N}_{i,j}),
\end{equation}
where $N_{i,j}$ and $\tilde{N}_{i,j}$ denote the predicted and ground truth normal maps, respectively and $\cdot$ indicates the dot product operation. When the predicted normal has a similar orientation as the ground truth, $N_{i,j} \cdot \tilde{N}_{i,j}$ approaches 1, the loss approaches 0, and vice versa.

\subsection{Masked lighting map}
\label{masked_light_map}

\begin{figure}[h]\centering
	\includegraphics[width=8cm]{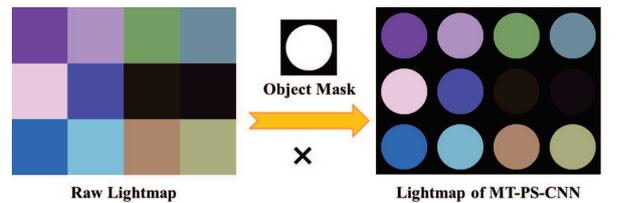}
	\caption{Illustration of generating masked lighting maps.}
	\label{fig:processlightmask}
\end{figure}

In the photometric stereo task, large field-of-view cameras are typically utilized to cover the entire target object for image capturing. As a result, a large area of input image covers invalid backgrounds and contains pixels of unchanged RGB values. Note that these background pixels present similar lighting variation patterns as the object pixels with insufficient reflectance observations (e.g., surfaces made of highly absorptive dark materials or with cast shadows) and will cause confusion for surface normal inference, particularly in intraframe spatial convolutions. Therefore, we argue that it is important to exclude such invalid background pixels in the training/testing of CNN-based surface normal estimation models.  

As a simple yet effective solution, we utilize the easily obtained object mask to extract pixels of the target object and eliminate invalid background regions. More specifically, we follow the conventional practice to replicate the 3-vector light direction of a captured image along the $x$ and $y$ directions to obtain a $3\times H \times W$ lighting map \cite{Chen2018ECCV_PS-FCN} and then apply the binary object mask (0 mask value defines a background pixel and 1 mask value defines an object pixel) to generate the masked lighting maps as illustrated in Fig.~\ref{fig:processlightmask}. In contrast to the existing uncalibrated photometric stereo method which concatenates object masks with individual captured images for light source estimation \cite{chen2020learned}, we refer to the binary object mask to eliminate adverse interference from invalid background regions and thus achieve more accurate surface normal estimation results. We will experimentally evaluate the effectiveness of utilizing masked lighting maps to improve the accuracy of normal estimation for surfaces made of highly absorptive dark materials or with specular highlights and cast shadows.

\section{Experimental Results}
\label{experiments}

In this section, we systematically evaluate the performance of our proposed MT-PS-CNN and compare it with the state-of-the-art photometric stereo methods on commonly used synthetic (MERL \cite{Matusik2003}) and real-world (DiLiGenT \cite{shi2016benchmark} and Light Stage Data Gallery \cite{chabert2006relighting}) datasets. We also validate the effectiveness of our proposed method using our own captured photometric stereo images of polished steel surfaces with tiny scratches.

\subsection{Implementation details}
All experiments were performed on a PC with GeForce GTX 1080Ti and 96 GB RAM. For training and testing, our model was implemented in PyTorch using 723 K learnable parameters. The Adam optimizer is used to optimize our networks with parameters $\beta_1=0.9$ and $\beta_2=0.999$. We use the \emph{Blobby} and \emph{Sculpture} datasets provided by \cite{Chen2018ECCV_PS-FCN} as our training data. There are 85212 samples on both datasets, where each sample contains 64 images rendered under randomly sampling light directions. The training process takes 30 epochs with a batch size of 32 (approximately 17 hours). We applied the same data augmentation technique as suggested in PS-FCN \cite{Chen2018ECCV_PS-FCN}, except adding extra noise disturbances. The accuracy of surface normal estimation is quantitatively evaluated by computing the mean angular error (MAE) between the predicted and ground truth normal maps as follows:
\begin{equation}
	{MAE}=\arccos{\left [\frac{1}{K}\sum_{k}(1-N_{k} \cdot \tilde{N}_{k})\right ] },
\end{equation}
where $N_{k}$ and $\tilde{N}_{k}$ denote the predicted and ground truth normal maps, respectively, and $K$ denotes the total number of target pixels in the normal map. Note that a lower MAE value indicates higher accuracy of normal estimation.

\subsection{Performance Analysis}

In this section, we establish ablation experiments to evaluate the effects of (1) the design of network architectures, (2) the incorporation of a masked lighting map, and (3) the setting of kernel sizes.

\subsubsection{Evaluation of network designs}
\label{network_design}

\begin{figure}[h]\centering
	\includegraphics[width=8cm]{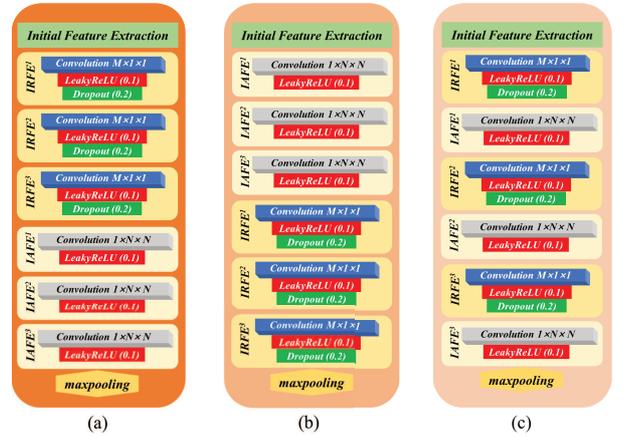}
	\caption{A number of network design alternatives for inter- and intraframe feature extraction. (a)T-IRFE-IAFE; (b)T-IAFE-IRFE; (c)M-IRFE-IAFE.}\label{fig:order}
\end{figure}

In this subsection, we discuss the best method to deploy interframe and intraframe feature extraction modules to achieve high-accuracy surface normal estimation and experimentally evaluate the performances of three network architectures as illustrated in Fig.~\ref{fig:order}.

The design of a CNN architecture to extract inter- and intraframe features for accurate surface normal estimation involves two critical design issues. The first issue is that both frame-to-frame (inter-) and per-frame (intra-) observations provide important information for the photometric stereo task and which feature extraction module should be deployed first. The second issue is whether the CNN architecture should be divided into two individual stages to perform inter- and intraframe feature extraction separately or mix the inter- and intraframe feature extraction steps.

Based on the two abovementioned critical design issues, we design three different network alternatives (Fig.~\ref{fig:order}) that employ different schemes to deploy inter- and intraframe feature extraction modules as follows: 

(1) Two-stage IRFE-IAFE design (T-IRFE-IAFE): The first design incorporates a two-stage architecture by deploying inter- and intraframe feature extraction steps in a cascaded manner. More specifically, it first deploys a number of IRFE blocks $\left(IRFE^{1}\Rightarrow IRFE^{2} \hdots \Rightarrow IRFE^{K}\right)$ to compute interframe features based on frame-to-frame lighting variations of individual pixels and then a number of IAFE blocks $\left(IAFE^{1}\Rightarrow IAFE^{2} \hdots \Rightarrow IAFE^{L}\right)$ to compute intraframe features by analysing intensity variation among spatial neighbouring pixels.

(2) Two-stage IAFE-IRFE design (T-IAFE-IRFE): The second design also utilizes a two-stage architecture but switches the order to deploy inter- and intraframe feature extraction modules. Thus, it first deploys a number of IAFE blocks $\left(IAFE^{1}\Rightarrow IAFE^{2} \hdots \Rightarrow IAFE^{L}\right)$ and then a number of IRFE blocks $\left(IRFE^{1}\Rightarrow IRFE^{2} \hdots \Rightarrow IRFE^{K}\right)$ to compute features for surface normal estimation.

(3) Mixed IRFE-IAFE design (M-IRFE-IAFE): Different from the above designs based on two-stage architecture, M-IRFE-IAFE deploys inter- and intraframe feature extraction steps in an alternative manner. More specifically, it makes use of a number of grouped IRFE and IAFE blocks $([IRFE^{1}, IAFE^{1}]\Rightarrow  [IRFE^{2}, IAFE^{2}]\hdots \Rightarrow [IRFE^{K},$ $ IAFE^{K}])$ to compute inter- and intraframe features in different convolutional stages.

In our implementation, we experimentally set $K=L=3$ to achieve a good balance between model complexity and good performance. 
For a fair comparison, these three different network design alternatives (T-IRFE-IAFE, T-IAFE-IRFE, and M-IRFE-IAFE) are trained and evaluated using the same parameters ($M=3$, $N=3$, $K=3$, and $L=3$). The comparative results (MAE) on the DiLiGenT dataset are shown in Table~\ref{tab:order}.

\begin{table}[htbp]
	\Huge
	\renewcommand{\arraystretch}{1.3}
	\caption{Quantitative evaluation (MAE) of three different architectures (T-IRFE-IAFE, T-IAFE-IRFE, and M-IRFE-IAFE) to deploy inter- and intraframe feature extraction blocks on the DiLiGenT dataset.}
	\centering
	
	\resizebox{\columnwidth}{!}{	
		\begin{tabular}{cccccccccccc}
			\hline
			Method & BALL  & CAT   & POT1  & BEAR  & POT2  & BUDD. & GOBL. & READ. & COW   & HARV. & Avg. \\
			\hline
			T-IR.-IA. & \cellcolor[rgb]{ .396,  .745,  .482}2.82  & \cellcolor[rgb]{ .533,  .788,  .51}\textbf{5.79}  & \cellcolor[rgb]{ .584,  .804,  .522}\textbf{6.92}  & \cellcolor[rgb]{ .525,  .788,  .51}\textbf{5.59}  & \cellcolor[rgb]{ .612,  .816,  .529}\textbf{7.55}  & \cellcolor[rgb]{ .584,  .804,  .522}\textbf{6.93}  & \cellcolor[rgb]{ .631,  .82,  .533}\textbf{7.98}  & \cellcolor[rgb]{ .812,  .875,  .569}11.85  & \cellcolor[rgb]{ .627,  .82,  .529}\textbf{7.82}  & \cellcolor[rgb]{ .918,  .91,  .592}\textbf{14.18}  & \cellcolor[rgb]{ .624,  .816,  .529}\textbf{7.74}  \\
			T-IA.-IR. & \cellcolor[rgb]{ .388,  .745,  .482}\textbf{2.58}  & \cellcolor[rgb]{ .545,  .792,  .514}6.02  & \cellcolor[rgb]{ .588,  .808,  .522}6.99  & \cellcolor[rgb]{ .573,  .8,  .518}6.62  & \cellcolor[rgb]{ .667,  .831,  .541}8.68  & \cellcolor[rgb]{ .604,  .812,  .525}7.34  & \cellcolor[rgb]{ .702,  .843,  .545}9.46  & \cellcolor[rgb]{ .894,  .902,  .588}13.68  & \cellcolor[rgb]{ .671,  .831,  .541}8.79  & \cellcolor[rgb]{ 1,  .937,  .612}15.95  & \cellcolor[rgb]{ .663,  .831,  .537}8.61  \\
			M-IR.-IA. & \cellcolor[rgb]{ .396,  .745,  .482}2.76  & \cellcolor[rgb]{ .541,  .792,  .514}5.99  & \cellcolor[rgb]{ .6,  .812,  .525}7.24  & \cellcolor[rgb]{ .58,  .804,  .522}6.84  & \cellcolor[rgb]{ .624,  .82,  .529}7.79  & \cellcolor[rgb]{ .6,  .812,  .525}7.23  & \cellcolor[rgb]{ .651,  .827,  .537}8.39  & \cellcolor[rgb]{ .804,  .875,  .569}\textbf{11.71}  & \cellcolor[rgb]{ .647,  .827,  .537}8.32  & \cellcolor[rgb]{ .945,  .918,  .6}14.79  & \cellcolor[rgb]{ .639,  .824,  .533}8.11  \\
			\hline
		\end{tabular}%
		\label{tab:order}
	}
\end{table}

It is observed that the design of T-IRFE-IAFE significantly outperforms the design of T-IAFE-IRFE, and the average MAE of 10 objects is reduced by $0.87^\circ$.  The comparative result shows that it is better to first perform interframe feature extraction and then intraframe feature extraction. This finding might be explained by the fact that although both inter- and intraframe features are useful for normal prediction, the interframe information is sensitive to outliers (specularity, shadow, etc.). Thus, it provides more important information to generate high-fidelity normal maps. In comparison, the intraframe features provide complementary information to efficiently eliminate the interference from outliers but adversely smooth out textured details. Our experimental results are consistent with previous research findings that pixelwise methods are generally more accurate for surface normal estimation than framewise methods if dense input images are provided \cite{Ikehata2018ECCV-CNN-PS, Santo2017}. Therefore, it is reasonable to first extract interframe features that provide fundamental information for the photometric stereo task and then perform spatial reasoning through intraframe feature extraction to further improve the accuracy of surface normal estimation.

Another interesting finding is that although M-IRFE-IAFE (altering inter- and intraframe feature extraction steps) is proven to be an effective design in many video sequence analysis tasks, such as gesture/action recognition \cite{Tran2018}, it is not very suitable for the photometric stereo task (M-IRFE-IAFE vs. T-IRFE-IAFE: $8.106^\circ$ vs. $7.741^\circ$). Given that inter- and intraframe features provide two  different types of information for estimating normal maps, it is more reasonable to divide the entire feature extraction process into two individual stages instead of mixing the inter- and intraframe feature extraction steps. 


\subsubsection{Effectiveness of masked lighting map}

\begin{table}[h]
	\renewcommand{\arraystretch}{1.3}
	\caption{Comparative results of various surface normal estimation models with/without referring to the masked lighting maps. For \emph{Sphere} and \emph{Bunny} objects, we calculate the average MAE values for 100 different materials.}
	\small
	\centering
	\label{table_1}
	\begin{tabular}{ccccc}
		\hline
		Object & Method & M=3, N=1 & M=1, N=3  & M=3, N=3 \\
		\hline
		\multirow{2}[0]{*}{\emph{Sphere}} & Mask & 8.89 & \textbf{4.51}    & \textbf{4.16}  \\
		& No-Mask  & \textbf{8.82}  & 5.97   & 5.73  \\
		\hline
		\multirow{2}[0]{*}{\emph{Bunny}} & Mask & 7.93  & \textbf{4.39}    & \textbf{4.03}  \\
		& No-Mask  & \textbf{7.86} & 5.01   & 4.73  \\
		\hline
	\end{tabular}%
	\label{tab:lightmask}%
\end{table}


In this subsection, we experimentally evaluate the effectiveness of utilizing masked lighting maps to improve the accuracy of normal estimation. Here, we adopt the best-performing two-stage CNN architecture (T-IRFE-IAFE) for inter- and intraframe feature extraction and set kernel size parameters $M$ (defining how many adjacent frames to process) and $N$ (defining how many neighbouring pixels to process) to different values. Note that the model will only perform interframe convolutions by setting $M=1$ given that it only considers intensity variation among spatial neighbouring pixels on the current frame. Similarly, the model will become a per-pixel method and compute interframe features exclusively based on lighting variations of single pixels by setting $N=1$.


\begin{figure}[h]\centering
	\includegraphics[width=8cm]{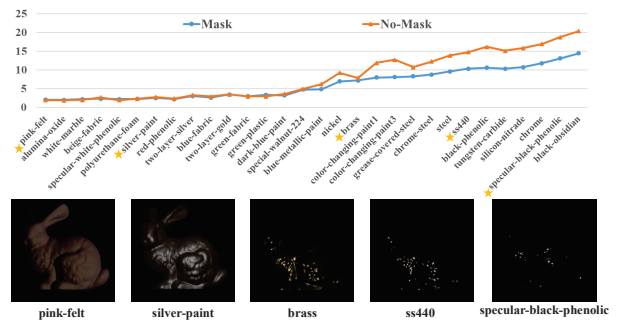}
	\caption{Quantitative comparison of CNN models ($M=3, N=3$) with/without referring to the object mask for the \emph{Bunny} object made of 30 different materials. Images in the second row present samples of 5 representative materials.}
	\label{fig:lightmask_exp1}
\end{figure}
To test our model on different materials, we use a synthetic dataset (\emph{Sphere} and \emph{Bunny} object), rendered with 100 different BRDFs. 
We calculate the average MAE values of various CNN models with/without referring to the masked lighting maps for 100 different materials, as illustrated in Table~\ref{tab:lightmask}. It is observed that the performance of the per-pixel CNN model ($M=3, N=1$) remains almost unchanged after incorporating the masked lighting map. In comparison, the CNN-based models performing intraframe spatial convolutions (setting $N=3$) achieved more accurate surface normal estimation results by referring to the binary object mask, which defines pixels of target and background. For instance, the average MAE of 100 different materials for \emph{Sphere} is significantly reduced from $5.97^\circ$ to $4.51^\circ$ for the model using $M=1, N=3$ and from $5.73^\circ$ to $4.16^\circ$ for the model using $M=3, N=3$. The quantitative evaluation results illustrate the importance of integrating the easily obtained object mask in CNN models that involve intraframe spatial convolutions, eliminating adverse interference from invalid background regions for high-accuracy surface normal estimation.

In Fig. \ref{fig:lightmask_exp1}, we show the calculated MAE values using CNN models ($M=3, N=3$) with/without referring to the object mask for \emph{Bunny} objects made of 30 representative materials. Note that the reflectance observations of background pixels remain zero, which presents similar variation patterns to those of object pixels with inadequate reflectance observations (e.g., made of highly absorptive dark materials). As a result, utilizing the binary object mask to eliminate adverse interference from invalid background regions leads to a significant increase in MAE values for objects of dark materials such as \emph{ss400} and \emph{specular-black-phenoli}, as shown in the right part of Fig. \ref{fig:lightmask_exp1}. In comparison, such improvement is almost neglectable for the \emph{Bunny} object made of light materials such as \emph{pink-felt} and \emph{silver-paint}, as illustrated in the left part of Fig. \ref{fig:lightmask_exp1}. 

\begin{figure}[h]\centering
	\Huge
	\includegraphics[width=8cm]{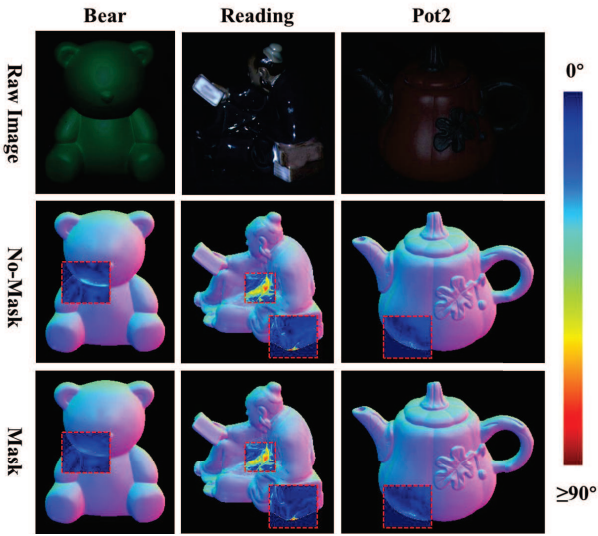}
	\caption{Qualitative results of CNN models with/without referring to the object mask for real-world objects (\emph{Bear}, \emph{Reading} and \emph{Pot2}) in the DiLiGenT dataset. More accurate normal estimation results are achieved for regions with cast shadows by referring to the easily obtained object mask.We purposely show error maps in the dashed boxes to better visualize normal estimation results in regions with complex geometry and obvious cast shadows.}
	\label{fig:mask_result_pic}
\end{figure}

In Fig. \ref{fig:mask_result_pic}, we visualize the surface normal estimation results with/without considering the object mask for real-world objects (\emph{Bear}, \emph{Reading} and \emph{Pot2}) in the DiLiGenT dataset. More accurate surface normal estimation results are generated in regions with complex geometries and obvious cast shadows by referring to the easily obtained object mask. 


\subsubsection{Setting kernel sizes}

\begin{table}[h]
	\Huge
	\renewcommand{\arraystretch}{1.3}
	\caption{The comparative results (MAE) of MT-PS-CNN models using different $M$ and $N$ parameters on objects in the DiLiGenT benchmark.}
	\centering
	\label{tab:Hyper-parameter experiment}
	\resizebox{\columnwidth}{!}{
		\begin{tabular}{c|c|ccccccccccc}
			\hline
			\multicolumn{2}{c}{Kernel Size}  & BALL  & CAT   & POT1  & BEAR  & POT2  & BUDD. & GOBL. & READ. & COW   & HARV. & Avg. \\
			\hline
			\multirow{4}[0]{*}{N=3} & M=1   & \cellcolor[rgb]{ .42,  .753,  .486}2.97  & \cellcolor[rgb]{ .569,  .8,  .518}6.23  & \cellcolor[rgb]{ .643,  .824,  .533}7.90  & \cellcolor[rgb]{ .608,  .812,  .525}7.13  & \cellcolor[rgb]{ .749,  .855,  .557}10.24  & \cellcolor[rgb]{ .659,  .827,  .537}8.23  & \cellcolor[rgb]{ .706,  .843,  .549}9.30  & \cellcolor[rgb]{ .89,  .902,  .588}13.37  & \cellcolor[rgb]{ .737,  .855,  .553}10.01  & \cellcolor[rgb]{ 1,  .937,  .612}15.80  & \cellcolor[rgb]{ .698,  .839,  .545}9.12  \\
			& M=3   & \cellcolor[rgb]{ .416,  .753,  .486}2.82  & \cellcolor[rgb]{ .553,  .796,  .518}5.95  & \cellcolor[rgb]{ .596,  .808,  .525}6.91  & \cellcolor[rgb]{ .533,  .788,  .51}5.46  & \cellcolor[rgb]{ .651,  .827,  .537}8.10  & \cellcolor[rgb]{ .604,  .812,  .525}7.05  & \cellcolor[rgb]{ .663,  .831,  .537}8.38  & \cellcolor[rgb]{ .82,  .878,  .573}\textbf{11.85}  & \cellcolor[rgb]{ .655,  .827,  .537}8.16  & \cellcolor[rgb]{ .922,  .91,  .592}14.09  & \cellcolor[rgb]{ .643,  .824,  .533}7.88  \\
			& M=5  & \cellcolor[rgb]{ .388,  .745,  .482}2.29  & \cellcolor[rgb]{ .553,  .796,  .514}5.87  & \cellcolor[rgb]{ .6,  .808,  .525}6.92  & \cellcolor[rgb]{ .549,  .792,  .514}5.79  & \cellcolor[rgb]{ .596,  .808,  .525}\textbf{6.89}  & \cellcolor[rgb]{ .596,  .808,  .525}\textbf{6.85}  & \cellcolor[rgb]{ .643,  .824,  .533}7.88  & \cellcolor[rgb]{ .824,  .882,  .573}11.94  & \cellcolor[rgb]{ .624,  .82,  .529}7.48  & \cellcolor[rgb]{ .906,  .906,  .588}13.71  & \cellcolor[rgb]{ .627,  .82,  .529}\textbf{7.56}  \\
			& M=7   & \cellcolor[rgb]{ .388,  .745,  .482}\textbf{2.21}  & \cellcolor[rgb]{ .533,  .788,  .51}\textbf{5.46}  & \cellcolor[rgb]{ .58,  .804,  .522}\textbf{6.53}  & \cellcolor[rgb]{ .541,  .792,  .514}5.62  & \cellcolor[rgb]{ .62,  .816,  .529}7.42  & \cellcolor[rgb]{ .6,  .812,  .525}6.99  & \cellcolor[rgb]{ .639,  .824,  .533}\textbf{7.86}  & \cellcolor[rgb]{ .851,  .89,  .58}12.56  & \cellcolor[rgb]{ .616,  .816,  .529}\textbf{7.30}  & \cellcolor[rgb]{ .914,  .91,  .592}13.93  & \cellcolor[rgb]{ .627,  .82,  .533}7.59  \\
			\hline
			\multirow{3}[0]{*}{M=5} & N=1   & \cellcolor[rgb]{ .412,  .753,  .486}2.80  & \cellcolor[rgb]{ .541,  .792,  .514}5.63  & \cellcolor[rgb]{ .592,  .808,  .525}6.74  & \cellcolor[rgb]{ .6,  .812,  .525}6.94  & \cellcolor[rgb]{ .702,  .843,  .549}9.23  & \cellcolor[rgb]{ .62,  .816,  .529}7.42  & \cellcolor[rgb]{ .749,  .859,  .557}10.28  & \cellcolor[rgb]{ .851,  .89,  .58}12.51  & \cellcolor[rgb]{ .843,  .886,  .576}12.32  & \cellcolor[rgb]{ .973,  .925,  .604}15.22  & \cellcolor[rgb]{ .686,  .839,  .545}8.91  \\
			& N=3   & \cellcolor[rgb]{ .388,  .745,  .482}2.29  & \cellcolor[rgb]{ .553,  .796,  .514}5.87  & \cellcolor[rgb]{ .6,  .808,  .525}6.92  & \cellcolor[rgb]{ .549,  .792,  .514}5.79  & \cellcolor[rgb]{ .596,  .808,  .525}6.89  & \cellcolor[rgb]{ .596,  .808,  .525}6.85  & \cellcolor[rgb]{ .643,  .824,  .533}7.88  & \cellcolor[rgb]{ .824,  .882,  .573}11.94  & \cellcolor[rgb]{ .624,  .82,  .529}7.48  & \cellcolor[rgb]{ .906,  .906,  .588}13.71  & \cellcolor[rgb]{ .627,  .82,  .529}7.56  \\
			& N=5   & \cellcolor[rgb]{ .392,  .745,  .482}2.30  & \cellcolor[rgb]{ .584,  .804,  .522}6.57  & \cellcolor[rgb]{ .608,  .812,  .529}7.16  & \cellcolor[rgb]{ .518,  .784,  .51}\textbf{5.11}  & \cellcolor[rgb]{ .647,  .824,  .533}7.97  & \cellcolor[rgb]{ .608,  .812,  .525}7.13  & \cellcolor[rgb]{ .659,  .827,  .537}8.24  & \cellcolor[rgb]{ .851,  .89,  .58}12.51  & \cellcolor[rgb]{ .624,  .816,  .529}7.45  & \cellcolor[rgb]{ .898,  .906,  .588}\textbf{13.60}  & \cellcolor[rgb]{ .639,  .824,  .533}7.80  \\
			\hline
		\end{tabular}%
	}
\end{table}

The kernel sizes of 3D convolutional layers ($M$ and $N$) in inter- and intraframe feature extractors are critical parameters to determine the complexity and performance of the proposed model. The performances of MT-PS-CNN models using different $M$ and $N$ parameters are experimentally evaluated, and the comparative results (MAE values) on objects in the DiLiGenT benchmark are shown in Table~\ref{tab:Hyper-parameter experiment}. It is observed that networks using larger kernel sizes generally produce lower MAE values. A more complex model (considering more adjacent frames and more neighbouring pixels) provides a better expressive/generalization ability to achieve more accurate surface normal estimation results. However, the improvements become insignificant when $M$ and $N$ are greater than 3. In our implementation, we set $M=N=3$ to achieve a good balance between model complexity and good performance.

\subsection{Comparisons with state-of-the-arts methods}

\begin{table}[h]
	
	\centering
	\renewcommand{\arraystretch}{1.3}
	\caption{Quantitative results of our proposed MT-PS-CNN model and state-of-the-art photometric stereo methods using the DiLiGenT benchmark dataset.}
	\resizebox{\columnwidth}{!}{
		\begin{threeparttable}
			\begin{tabular}{cccccccccccc}
				\hline
				Method & BALL  & CAT   & POT1  & BEAR  & POT2  & BUDD. & GOBL. & READ. & COW   & HARV. & Avg. \\
				\hline
				Proposed & \cellcolor[rgb]{ .404,  .749,  .482}2.29  & \cellcolor[rgb]{ .478,  .773,  .498}5.87  & \cellcolor[rgb]{ .502,  .78,  .506}6.92  & \cellcolor[rgb]{ .478,  .773,  .498}5.79  & \cellcolor[rgb]{ .502,  .78,  .506}6.89  & \cellcolor[rgb]{ .498,  .78,  .506}\textbf{6.85}  & \cellcolor[rgb]{ .522,  .784,  .51}7.88  & \cellcolor[rgb]{ .608,  .812,  .525}11.94  & \cellcolor[rgb]{ .514,  .784,  .506}7.48  & \cellcolor[rgb]{ .643,  .824,  .533}\textbf{13.71}  & \cellcolor[rgb]{ .514,  .784,  .506}\textbf{7.56}  \\
				JU-19\cite{Li2019} & \cellcolor[rgb]{ .404,  .749,  .486}2.40  & \cellcolor[rgb]{ .482,  .773,  .502}6.11  & \cellcolor[rgb]{ .494,  .776,  .502}6.54  & \cellcolor[rgb]{ .467,  .769,  .498}\textbf{5.23}  & \cellcolor[rgb]{ .514,  .784,  .506}7.49  & \cellcolor[rgb]{ .565,  .8,  .518}9.89  & \cellcolor[rgb]{ .537,  .792,  .514}8.61  & \cellcolor[rgb]{ .643,  .824,  .533}13.68  & \cellcolor[rgb]{ .522,  .784,  .51}7.98  & \cellcolor[rgb]{ .694,  .839,  .545}16.18  & \cellcolor[rgb]{ .533,  .788,  .51}8.41  \\
				CH-18\cite{Chen2018ECCV_PS-FCN} & \cellcolor[rgb]{ .416,  .753,  .486}2.82  & \cellcolor[rgb]{ .486,  .773,  .502}6.16  & \cellcolor[rgb]{ .506,  .78,  .506}7.13  & \cellcolor[rgb]{ .514,  .784,  .506}7.55  & \cellcolor[rgb]{ .506,  .78,  .506}7.25  & \cellcolor[rgb]{ .522,  .784,  .51}7.91  & \cellcolor[rgb]{ .537,  .788,  .514}8.60  & \cellcolor[rgb]{ .635,  .82,  .533}13.33  & \cellcolor[rgb]{ .51,  .78,  .506}7.33  & \cellcolor[rgb]{ .686,  .839,  .545}15.85  & \cellcolor[rgb]{ .533,  .788,  .51}8.39  \\
				SI-18$^*$\cite{Ikehata2018ECCV-CNN-PS} & \cellcolor[rgb]{ .4,  .749,  .482}2.20  & \cellcolor[rgb]{ .451,  .765,  .494}\textbf{4.60}  & \cellcolor[rgb]{ .471,  .769,  .498}\textbf{5.40}  & \cellcolor[rgb]{ .612,  .816,  .529}12.30  & \cellcolor[rgb]{ .482,  .773,  .502}\textbf{6.00}  & \cellcolor[rgb]{ .522,  .784,  .51}7.90  & \cellcolor[rgb]{ .51,  .78,  .506}\textbf{7.30}  & \cellcolor[rgb]{ .62,  .816,  .529}12.60  & \cellcolor[rgb]{ .522,  .784,  .51}7.90  & \cellcolor[rgb]{ .647,  .824,  .537}13.90  & \cellcolor[rgb]{ .522,  .784,  .51}8.01  \\
				TM-18\cite{Taniai2018} & \cellcolor[rgb]{ .388,  .745,  .482}\textbf{1.47}  & \cellcolor[rgb]{ .471,  .769,  .498}5.44  & \cellcolor[rgb]{ .482,  .773,  .502}6.09  & \cellcolor[rgb]{ .478,  .773,  .498}5.79  & \cellcolor[rgb]{ .518,  .784,  .51}7.76  & \cellcolor[rgb]{ .573,  .8,  .522}10.36  & \cellcolor[rgb]{ .596,  .808,  .525}11.47  & \cellcolor[rgb]{ .588,  .808,  .522}\textbf{11.03}  & \cellcolor[rgb]{ .486,  .776,  .502}\textbf{6.32}  & \cellcolor[rgb]{ .831,  .882,  .573}22.59  & \cellcolor[rgb]{ .541,  .792,  .514}8.83  \\
				HI-17\cite{Santo2017} & \cellcolor[rgb]{ .396,  .745,  .482}2.02  & \cellcolor[rgb]{ .494,  .776,  .502}6.54  & \cellcolor[rgb]{ .502,  .78,  .506}7.05  & \cellcolor[rgb]{ .486,  .776,  .502}6.31  & \cellcolor[rgb]{ .522,  .784,  .51}7.86  & \cellcolor[rgb]{ .62,  .816,  .529}12.68  & \cellcolor[rgb]{ .592,  .808,  .525}11.28  & \cellcolor[rgb]{ .682,  .835,  .541}15.51  & \cellcolor[rgb]{ .522,  .784,  .51}8.01  & \cellcolor[rgb]{ .71,  .843,  .549}16.86  & \cellcolor[rgb]{ .553,  .796,  .514}9.41  \\
				ST-14\cite{Shi2014} & \cellcolor[rgb]{ .392,  .745,  .482}1.74  & \cellcolor[rgb]{ .482,  .773,  .502}6.12  & \cellcolor[rgb]{ .49,  .776,  .502}6.51  & \cellcolor[rgb]{ .482,  .773,  .502}6.12  & \cellcolor[rgb]{ .541,  .792,  .514}8.78  & \cellcolor[rgb]{ .576,  .804,  .522}10.60  & \cellcolor[rgb]{ .569,  .8,  .518}10.09  & \cellcolor[rgb]{ .643,  .824,  .533}13.63  & \cellcolor[rgb]{ .647,  .824,  .537}13.93  & \cellcolor[rgb]{ .89,  .902,  .588}25.44  & \cellcolor[rgb]{ .573,  .8,  .518}10.30  \\
				IA-14\cite{Ikehata2014} & \cellcolor[rgb]{ .427,  .757,  .49}3.34  & \cellcolor[rgb]{ .498,  .776,  .502}6.74  & \cellcolor[rgb]{ .494,  .776,  .502}6.64  & \cellcolor[rgb]{ .506,  .78,  .506}7.11  & \cellcolor[rgb]{ .541,  .792,  .514}8.77  & \cellcolor[rgb]{ .576,  .804,  .522}10.47  & \cellcolor[rgb]{ .561,  .796,  .518}9.71  & \cellcolor[rgb]{ .655,  .827,  .537}14.19  & \cellcolor[rgb]{ .627,  .82,  .533}13.05  & \cellcolor[rgb]{ .902,  .906,  .588}25.95  & \cellcolor[rgb]{ .576,  .804,  .522}10.60  \\
				BASELINE\cite{Woodham1980} & \cellcolor[rgb]{ .443,  .761,  .49}4.10  & \cellcolor[rgb]{ .533,  .788,  .51}8.41  & \cellcolor[rgb]{ .541,  .792,  .514}8.89  & \cellcolor[rgb]{ .533,  .788,  .51}8.39  & \cellcolor[rgb]{ .663,  .831,  .537}14.65  & \cellcolor[rgb]{ .667,  .831,  .541}14.92  & \cellcolor[rgb]{ .745,  .855,  .557}18.50  & \cellcolor[rgb]{ .773,  .863,  .561}19.80  & \cellcolor[rgb]{ .894,  .902,  .588}25.60  & \cellcolor[rgb]{ 1,  .937,  .612}30.62  & \cellcolor[rgb]{ .678,  .835,  .541}15.39  \\
				\hline
			\end{tabular}%
			
			\begin{tablenotes}
				\item $^*$ indicates that we use all 96 images to estimate the normal map of \emph{Bear}. The result shown in SI-18 \cite{Ikehata2018ECCV-CNN-PS} (\emph{Bear} 4.1) was achieved by discarding the first 20 input images.
			\end{tablenotes}
	\end{threeparttable}}
	\label{tab:Diligent}
\end{table}

In this subsection, we compared our proposed MT-PS-CNN model with a number of state-of-the-art photometric stereo solutions including IK-12\cite{Ikehata2012}, IA-14\cite{Ikehata2014}, ST-14\cite{Shi2014}, HI-17\cite{Santo2017}, TM-18\cite{Taniai2018}, CH-18\cite{Chen2018ECCV_PS-FCN}, SI-18\cite{Ikehata2018ECCV-CNN-PS}, and JU-19\cite{Li2019}. The source codes or evaluation results of these methods are publicly available.

\begin{figure}[htbp]\centering
	\includegraphics[width=8cm]{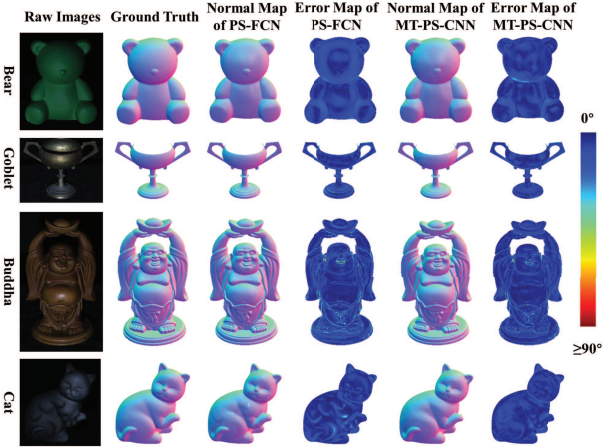}
	\caption{Surface normal estimation results using 96 input images on the DiLiGenT dataset. Zoom in to assess estimation results in regions with complex geometry.}
	\label{fig:dilgent}
\end{figure}

Quantitative evaluation results using the DiLiGenT benchmark dataset are provided in Table~\ref{tab:Diligent}. When the number of input
images is 96, our proposed MT-PS-CNN model achieved the lowest average MAE ($7.56^\circ$) on 10 real-world objects of DiLiGenT. It is worth mentioning that such improvements are particularly obvious for objects with complex geometry (e.g., \emph{Buddha}, \emph{Reading}, and \emph{Harvest}). 

\begin{figure}[h]\centering
	\includegraphics[width=0.9\linewidth]{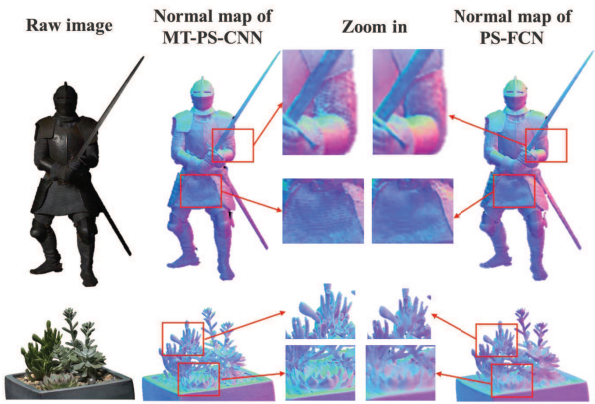}
	\caption{Surface normal estimation results using 36 input images on the Light Stage Data Gallery dataset. The zoomed in image provides more details of the highlighted regions.}
	\label{fig:Gallery}
\end{figure}

Figs.~\ref{fig:dilgent} and \ref{fig:Gallery} show some qualitative results on two real-world DiLiGenT and Light Stage Data Gallery datasets, respectively. Our proposed MT-PS-CNN model is capable of predicting a surface normal map with rich details. Moreover, it can generate more accurate surface normal estimation results in the highlighted regions with complex geometry compared with the state-of-the-art PS-FCN method (CH-18) \cite{Chen2018ECCV_PS-FCN}.


\begin{table}[h]
	\centering
	\small
	\renewcommand{\arraystretch}{1.3}
	\caption{Parameter number and running time of a number of state-of-the-art deep learning-based photometric stereo approaches.}
	
	\begin{tabular}{cccc}
		\hline
		
		& SI-18\cite{Ikehata2018ECCV-CNN-PS} & CH-18\cite{Chen2018ECCV_PS-FCN} & MT-PS-CNN \\
		\hline
		Parameter number & 2.65M & 2.21M & \textbf{0.72M} \\
		Running time (s) & 19.1  & 2.1   & \textbf{0.27} \\
		\hline
		
	\end{tabular}%
	\label{tab:parameter}
	
\end{table}

\begin{figure*}[t]\centering
	\includegraphics[width=17cm]{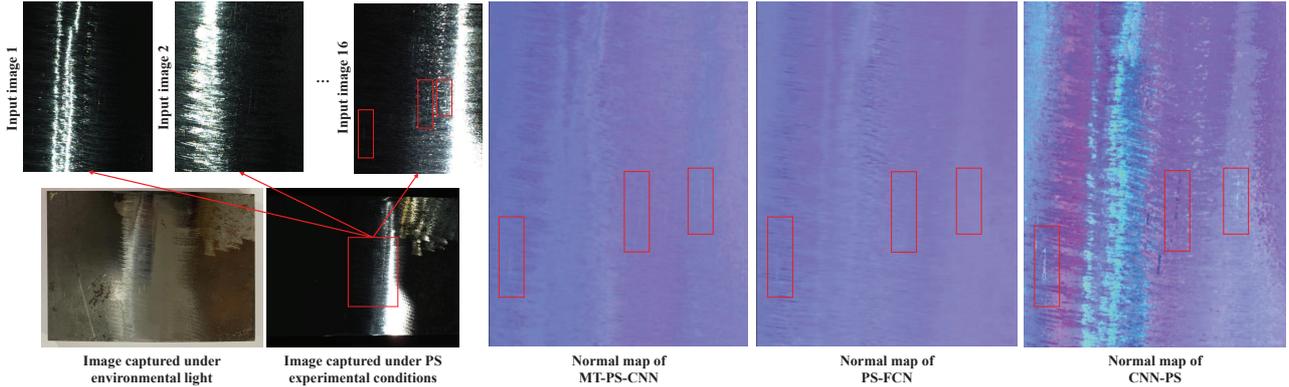}
	\caption{Normal map comparison of different methods on polished steel surfaces.}
	\label{fig:steel_surface}
\end{figure*}

Table~\ref{tab:parameter} shows the parameter number and running time of MT-PS-CNN and two representative deep learning-based photometric stereo approaches on a PC with GeForce GTX 1080Ti and 96 GB RAM. We repeat the estimation process 10 times and compute the average running time (the forward time of the network). For a fair comparison, different methods are applied for the normal estimation of the croped \emph{BallPNG} in DiLiGenT, and the input number is 96. Following the setting in SI-18 \cite{Ikehata2018ECCV-CNN-PS}, we set the number of different rotations for the rotational pseudoinvariance to 1. Our proposed MT-PS-CNN model achieves the highest normal estimation accuracy using significantly fewer parameters and runs much faster. Thus, our model is more suitable for real-time implementation in memory-restricted devices.

Another noticeable advantage of the proposed MT-PS-CNN is that it performs robustly when the number of input images significantly changes. In Table~\ref{tab:sparse Inputs}, we show the normal estimation results (the average MAE of 10 objects in the DiLiGenT dataset) of different deep-learning-based photometric stereo methods using 96, 16, 10, 8 and 6 input images. Note that JU-19\cite{Li2019} is designed to decrease the demands on the number of images for the photometric stereo task by learning the most informative images under different illumination conditions; thus, it performs the best based on 6 and 8 input images. However, its performance is suboptimal when processing dense input images. In comparison, our proposed method performs consistently well based on images captured in both sparse and dense lighting distributions. As illustrated in Table~\ref{tab:sparse Inputs detail}, MT-PS-CNN performs significantly better than other alternatives for objects with complex geometry (e.g., \emph{Buddha}, \emph{Reading}, and \emph{Harvest}), which represents a more challenging normal estimation task. 

\begin{table}[htbp]
	\renewcommand{\arraystretch}{1.3}
	\caption{Surface normal estimation results using different numbers of input images (96, 16, 10, 8, 6) on the DiLiGenT dataset. We calculate the average MAE for 10 objects.}
	\centering
	\small
	\begin{tabular}{cccccc}
		\hline
		Method & 96    & 16    & 10    & 8     & 6 \\
		\hline
		Proposed & \textbf{7.56 } & \textbf{8.82 } & \textbf{9.84 } & 10.75  & 12.30  \\
		JU-19\cite{Li2019} & 8.41  & 9.66  & 10.02  & \textbf{10.39 } & \textbf{12.16 } \\
		CH-18\cite{Chen2018ECCV_PS-FCN} & 8.39  & 9.37  & 10.33  & 11.13  & 12.56  \\
		\hline
	\end{tabular}%
	\label{tab:sparse Inputs}	
\end{table}

\begin{table}[h]
	\Huge
	\renewcommand{\arraystretch}{1.3}
	\caption{Surface normal estimation results of objects in the DiLiGenT dataset using 10 illumination directions. We randomly selected 10 images from 96 inputs and calculated the average MAE over 10 trials.}
	\centering
	\label{table_1}
	\resizebox{\columnwidth}{!}{
		\begin{tabular}{cccccccccccc}
			\hline
			Method & BALL  & CAT   & POT1  & BEAR  & POT2  & BUDD. & GOBL. & READ. & COW   & HARV. & Avg. \\
			\hline
			Proposed & \cellcolor[rgb]{ .392,  .745,  .482}4.20  & \cellcolor[rgb]{ .49,  .776,  .502}7.30  & \cellcolor[rgb]{ .533,  .788,  .51}8.78  & \cellcolor[rgb]{ .529,  .788,  .51}8.59  & \cellcolor[rgb]{ .569,  .8,  .518}9.85  & \cellcolor[rgb]{ .518,  .784,  .51}\textbf{8.25}  & \cellcolor[rgb]{ .584,  .804,  .522}\textbf{10.44}  & \cellcolor[rgb]{ .671,  .831,  .541}\textbf{13.17}  & \cellcolor[rgb]{ .596,  .808,  .525}10.84  & \cellcolor[rgb]{ .784,  .867,  .565}\textbf{16.97}  & \cellcolor[rgb]{ .565,  .8,  .518}\textbf{9.84}  \\
			JU-19\cite{Li2019} & \cellcolor[rgb]{ .388,  .745,  .482}\textbf{3.97}  & \cellcolor[rgb]{ .471,  .769,  .498}\textbf{6.70}  & \cellcolor[rgb]{ .49,  .776,  .502}\textbf{7.30}  & \cellcolor[rgb]{ .533,  .788,  .51}8.73  & \cellcolor[rgb]{ .565,  .8,  .518}9.74  & \cellcolor[rgb]{ .612,  .816,  .529}11.36  & \cellcolor[rgb]{ .584,  .804,  .522}10.46  & \cellcolor[rgb]{ .706,  .843,  .549}14.37  & \cellcolor[rgb]{ .576,  .804,  .522}10.19  & \cellcolor[rgb]{ .796,  .871,  .569}17.33  & \cellcolor[rgb]{ .573,  .8,  .522}10.02 \\
			CH-18 & \cellcolor[rgb]{ .396,  .745,  .482}4.26  & \cellcolor[rgb]{ .514,  .784,  .506}8.12  & \cellcolor[rgb]{ .565,  .8,  .518}9.84  & \cellcolor[rgb]{ .514,  .784,  .506}\textbf{8.07}  & \cellcolor[rgb]{ .549,  .796,  .514}\textbf{9.29}  & \cellcolor[rgb]{ .549,  .792,  .514}9.24  & \cellcolor[rgb]{ .592,  .808,  .525}10.61  & \cellcolor[rgb]{ .729,  .851,  .553}15.11  & \cellcolor[rgb]{ .569,  .8,  .518}\textbf{9.90}  & \cellcolor[rgb]{ .843,  .886,  .576}18.90  & \cellcolor[rgb]{ .58,  .804,  .522}10.33  \\
			\hline
	\end{tabular}}%
	\label{tab:sparse Inputs detail}
\end{table}

We also validate the effectiveness of our proposed MT-PS-CNN model on images captured using our own photometric stereo data acquisition system.  Fig.~\ref{fig:steel_surface} shows the comparative performance of PS-FCN \cite{Chen2018ECCV_PS-FCN}, CNN-PS \cite{Ikehata2018ECCV-CNN-PS} and the proposed MT-PS-CNN. It is observed that the PS-FCN model utilizes intraframe information to effectively suppress interference of specular highlights. However, some important details (e.g., scratches on the steel surface) are adversely smoothed out, as shown in Fig.~\ref{fig:steel_surface}.  In comparison, the CNN-PS model generates a normal map with rich textures/details, but its performance is significantly affected by specular highlights due to the lack of consideration of spatial information. Our proposed MT-PS-CNN model utilizes both inter- and intraframe information and thus can combine the advantages of the above two methods.

\section{Conclusion}
\label{conclusion}
In this paper, we present a complete photometric stereo acquisition/processing framework including a multilight source illumination and acquisition system and a two-stage CNN architecture to extract inter- and intraframe feature representations for high-quality surface normal estimation of non-Lambertian objects. We experimentally investigate a number of network design alternatives to identify the optimal scheme (T-IRFE-IAFE) to deploy interframe and intraframe feature extraction modules for the photometric stereo problem. Moreover, we integrate the easily obtained object mask in intraframe spatial convolutions to improve the accuracy of normal estimation for surfaces made of highly absorptive dark materials or with obvious cast shadows. The advantages of the proposed MT-PS-CNN model include producing more accurate normal estimation results using significantly few parameters and performing robustly under both dense and sparse image capturing configurations. The source code of the MT-PS-CNN model will be made publicly available.

We experimentally observed that the order of input images will affect the performance of our method based on inter- and intraframe 3D convolution. In the future, we plan to systematically investigate the optimal strategy to process image sequences to achieve high-accuracy and robust surface normal estimation. Moreover, our proposed MT-PS-CNN is a lightweight model, and it is possible to stack more convolutional layers/modules or adopt more complex network architectures (e.g., PS-FCN$^+$$^N$\cite{chen2020deep}, PX-NET\cite{logothetis2020px}) to further improve its accuracy.

\bibliographystyle{unsrt}
\bibliography{mybibfile}

\begin{thebibliography}{10}

\bibitem{song2018photometric}
Zhao Song, Ying Nie, and Zhan Song.
\newblock Photometric stereo with quasi-point light source.
\newblock {\em Optics and Lasers in Engineering}, 111:172--182, 2018.

\bibitem{villa2004surface}
Jes{\'u}s Villa and Juan~B Hurtado-Ramos.
\newblock Surface shape estimation from photometric images.
\newblock {\em Optics and lasers in engineering}, 42(4):461--468, 2004.

\bibitem{3dDetection}
Y.~{Fang}, G.~{Ding}, W.~{Wen}, F.~{Yuan}, Y.~{Yang}, Z.~{Fang}, and W.~{Lin}.
\newblock Salient object detection by spatiotemporal and semantic features in
  real-time video processing systems.
\newblock {\em IEEE Transactions on Industrial Electronics}, 67(11):9893--9903,
  2020.

\bibitem{ma2019calibration}
Long Ma, Jirui Liu, Xin Pei, Yanmin Hu, and Fengming Sun.
\newblock Calibration of position and orientation for point light source
  synchronously with single image in photometric stereo.
\newblock {\em Optics express}, 27(4):4024--4033, 2019.

\bibitem{xie2015practical}
Limin Xie, Zhan Song, Guohua Jiao, Xinhan Huang, and Kui Jia.
\newblock A practical means for calibrating an led-based photometric stereo
  system.
\newblock {\em Optics and Lasers in Engineering}, 64:42--50, 2015.

\bibitem{zhou2019digital}
Haowen Zhou, Xiaomeng Sui, Liangcai Cao, and Partha~P Banerjee.
\newblock Digital correlation of computer-generated holograms for 3d face
  recognition.
\newblock {\em Applied optics}, 58(34):G177--G186, 2019.

\bibitem{zhou20193}
Pei Zhou, Jiangping Zhu, and Zhisheng You.
\newblock 3-d face registration solution with speckle encoding based
  spatial-temporal logical correlation algorithm.
\newblock {\em Optics express}, 27(15):21004--21019, 2019.

\bibitem{Woodham1980}
Robert~J. Woodham.
\newblock {Photometric Method For Determining Surface Orientation From Multiple
  Images}.
\newblock {\em Optical Engineering}, 19(1):139--144, feb 1980.

\bibitem{Goldman2005}
D.B. Goldman, Brian Curless, Aaron Hertzmann, and S.M. Seitz.
\newblock {Shape and spatially-varying BRDFs from photometric stereo}.
\newblock In {\em Tenth IEEE International Conference on Computer Vision
  (ICCV'05) Volume 1}, volume~I, pages 341--348 Vol. 1. IEEE, 2005.

\bibitem{Shi2014}
Boxin Shi, Ping Tan, Yasuyuki Matsushita, and Katsushi Ikeuchi.
\newblock {Bi-Polynomial Modeling of Low-Frequency Reflectances}.
\newblock {\em IEEE Transactions on Pattern Analysis and Machine Intelligence},
  36(6):1078--1091, jun 2014.

\bibitem{Ikehata2018ECCV-CNN-PS}
Satoshi Ikehata.
\newblock {CNN-PS: CNN-based Photometric Stereo for General Non-Convex
  Surfaces}.
\newblock In {\em Proceedings of the European conference on computer vision
  (ECCV)}, pages 614--629, 2018.

\bibitem{Chen2018ECCV_PS-FCN}
Guanying Chen, Kai Han, and Kwan-Yee~K. Wong.
\newblock {PS-FCN: A Flexible Learning Framework for Photometric Stereo}.
\newblock In {\em Proceedings of the European conference on computer vision
  (ECCV)}, pages 3--19. jul 2018.

\bibitem{Chen2019CVPR}
Guanying Chen, Kai Han, Boxin Shi, Yasuyuki Matsushita, and Kwan Yee~K.K. Wong.
\newblock {Self-calibrating deep photometric stereo networks}.
\newblock In {\em Proceedings of the IEEE/CVF Conference on Computer Vision and
  Pattern Recognition (CVPR)}, pages 8731--8739, 2019.

\bibitem{Ackermann2015}
Jens Ackermann and Michael Goesele.
\newblock {A Survey of Photometric Stereo Techniques}.
\newblock {\em Foundations and Trends{\textregistered} in Computer Graphics and
  Vision}, 9(3-4):149--254, 2015.

\bibitem{Solomon1992}
Fredric Solomon and Katsushi Ikeuchi.
\newblock {Extracting the shape and roughness of specular lobe objects using
  four light photometric stereo}.
\newblock In {\em Proceedings 1992 IEEE Computer Society Conference on Computer
  Vision and Pattern Recognition}, number 7597, pages 466--471. IEEE Comput.
  Soc. Press.

\bibitem{Coleman1982}
E.~North Coleman and Ramesh Jain.
\newblock {Obtaining 3-dimensional shape of textured and specular surfaces
  using four-source photometry}.
\newblock {\em Computer Graphics and Image Processing}, 18(4):309--328, apr
  1982.

\bibitem{Wu2011}
Lun Wu, Arvind Ganesh, Boxin Shi, Yasuyuki Matsushita, Yongtian Wang, and
  Yi~Ma.
\newblock {Robust Photometric Stereo via Low-Rank Matrix Completion and
  Recovery}.
\newblock In {\em Lecture Notes in Computer Science (including subseries
  Lecture Notes in Artificial Intelligence and Lecture Notes in
  Bioinformatics)}, pages 703--717. 2011.

\bibitem{mukaigawa2007analysis}
Yasuhiro Mukaigawa, Yasunori Ishii, and Takeshi Shakunaga.
\newblock Analysis of photometric factors based on photometric linearization.
\newblock {\em JOSA A}, 24(10):3326--3334, 2007.

\bibitem{Miyazaki2010}
Daisuke Miyazaki, Kenji Hara, and Katsushi Ikeuchi.
\newblock {Median Photometric Stereo as Applied to the Segonko Tumulus and
  Museum Objects}.
\newblock {\em International Journal of Computer Vision}, 86(2-3):229--242, jan
  2010.

\bibitem{Torrance1967}
K.~E. Torrance and E.~M. Sparrow.
\newblock {Theory for Off-Specular Reflection From Roughened Surfaces*}.
\newblock {\em Journal of the Optical Society of America}, 57(9):1105, sep
  1967.

\bibitem{Kay1995}
Greg Kay and Terry Caelli.
\newblock {Estimating the parameters of an illumination model using photometric
  stereo}, 1995.

\bibitem{Georghiades2003}
Georghiades.
\newblock {Incorporating the Torrance and Sparrow model of reflectance in
  uncalibrated photometric stereo}.
\newblock In {\em Proceedings Ninth IEEE International Conference on Computer
  Vision}, volume~2, pages 816--823 vol.2. IEEE, 2003.

\bibitem{Cook1981}
Robert~L. Cook and Kenneth~E. Torrance.
\newblock {A reflectance model for computer graphics}.
\newblock In {\em Proceedings of the 8th annual conference on Computer graphics
  and interactive techniques - SIGGRAPH '81}, volume~15, pages 307--316, New
  York, New York, USA, 1981. ACM Press.

\bibitem{Hin-ShunChung2008}
{Hin-Shun Chung} and {Jiaya Jia}.
\newblock {Efficient photometric stereo on glossy surfaces with wide specular
  lobes}.
\newblock In {\em 2008 IEEE Conference on Computer Vision and Pattern
  Recognition}, pages 1--8. IEEE, jun 2008.

\bibitem{Goldman2010}
{Shape and Spatially-Varying BRDFs from Photometric Stereo}.
\newblock {\em IEEE Transactions on Pattern Analysis and Machine Intelligence},
  32(6):1060--1071, jun 2010.

\bibitem{Phong1975}
Bui~Tuong Phong.
\newblock {Illumination for computer generated pictures}.
\newblock {\em Communications of the ACM}, 18(6):311--317, jun 1975.

\bibitem{Blinn1977}
James~F. Blinn.
\newblock {Models of light reflection for computer synthesized pictures}.
\newblock In {\em Proceedings of the 4th annual conference on Computer graphics
  and interactive techniques - SIGGRAPH '77}, pages 192--198, New York, New
  York, USA, 1977. ACM Press.

\bibitem{ngan2005experimental}
Addy Ngan, Fr{\'e}do Durand, and Wojciech Matusik.
\newblock Experimental analysis of brdf models.
\newblock {\em Rendering Techniques}, 2005(16th):2, 2005.

\bibitem{stark2005barycentric}
Michael~M Stark, James Arvo, and Brian Smits.
\newblock Barycentric parameterizations for isotropic brdfs.
\newblock {\em IEEE transactions on visualization and computer graphics},
  11(2):126--138, 2005.

\bibitem{hertzmann2005example}
Aaron Hertzmann and Steven~M Seitz.
\newblock Example-based photometric stereo: Shape reconstruction with general,
  varying brdfs.
\newblock {\em IEEE Transactions on Pattern Analysis and Machine Intelligence},
  27(8):1254--1264, 2005.

\bibitem{Santo2017}
Hiroaki Santo, Masaki Samejima, Yusuke Sugano, Boxin Shi, and Yasuyuki
  Matsushita.
\newblock {Deep Photometric Stereo Network}.
\newblock In {\em 2017 IEEE International Conference on Computer Vision
  Workshops (ICCVW)}, pages 501--509. IEEE, 2017.

\bibitem{Li2019}
Junxuan Li, Antonio Robles-Kelly, Shaodi You, and Yasuyuki Matsushita.
\newblock {Learning to Minify Photometric Stereo}.
\newblock In {\em 2019 IEEE/CVF Conference on Computer Vision and Pattern
  Recognition (CVPR)}, volume 2019-June, pages 7560--7568. IEEE, jun 2019.

\bibitem{Zheng2019_SPLINE-Net}
Qian Zheng, Yiming Jia, Boxin Shi, Xudong Jiang, Ling-Yu Duan, and Alex Kot.
\newblock {SPLINE-Net: Sparse Photometric Stereo Through Lighting Interpolation
  and Normal Estimation Networks}.
\newblock In {\em 2019 IEEE/CVF International Conference on Computer Vision
  (ICCV)}, pages 8548--8557. IEEE, oct 2019.

\bibitem{FuLin2020}
Xi~Wang, Zhenxiong Jian, and Mingjun Ren.
\newblock {Non-Lambertian Photometric Stereo Network Based on Inverse
  Reflectance Model With Collocated Light}.
\newblock {\em IEEE Transactions on Image Processing}, 29(5):6032--6042, 2020.

\bibitem{Mukaigawa2007}
Yasuhiro Mukaigawa, Yasunori Ishii, and Takeshi Shakunaga.
\newblock {Analysis of photometric factors based on photometric linearization}.
\newblock {\em Journal of the Optical Society of America A}, 24(10):3326, oct
  2007.

\bibitem{Li2019CVPR}
Junxuan Li, Antonio Robles-Kelly, Shaodi You, and Yasuyuki Matsushita.
\newblock {Learning to Minify Photometric Stereo}.
\newblock In {\em 2019 IEEE/CVF Conference on Computer Vision and Pattern
  Recognition (CVPR)}, volume 2019-June, pages 7560--7568. IEEE, jun 2019.

\bibitem{Wang2020TIP}
Xi~Wang, ZhenXiong Jian, and Mingjun Ren.
\newblock {Non-Lambertian Photometric Stereo Network Based on Inverse
  Reflectance Model With Collocated Light}.
\newblock {\em IEEE Transactions on Image Processing}, 29(5):6032--6042, 2020.

\bibitem{Hartmann2017learned}
Wilfried Hartmann, Silvano Galliani, Michal Havlena, Luc Van~Gool, and Konrad
  Schindler.
\newblock Learned multi-patch similarity.
\newblock In {\em Proceedings of the IEEE International Conference on Computer
  Vision}, pages 1586--1594, 2017.

\bibitem{Wiles2017silnet}
Olivia Wiles and Andrew Zisserman.
\newblock Silnet: Single-and multi-view reconstruction by learning from
  silhouettes.
\newblock In {\em BMVC}, 2017.

\bibitem{chen2020learned}
Guanying Chen, Michael Waechter, Boxin Shi, Kwan-Yee~K Wong, and Yasuyuki
  Matsushita.
\newblock What is learned in deep uncalibrated photometric stereo?
\newblock In {\em European Conference on Computer Vision}, pages 745--762.
  Springer, 2020.

\bibitem{Matusik2003}
Wojciech Matusik, Hanspeter Pfister, Matt Brand, and Leonard McMillan.
\newblock {A data-driven reflectance model}.
\newblock {\em ACM Transactions on Graphics}, 22(3):759, jul 2003.

\bibitem{shi2016benchmark}
Boxin Shi, Zhe Wu, Zhipeng Mo, Dinglong Duan, Sai-Kit Yeung, and Ping Tan.
\newblock A benchmark dataset and evaluation for non-lambertian and
  uncalibrated photometric stereo.
\newblock In {\em Proceedings of the IEEE Conference on Computer Vision and
  Pattern Recognition}, pages 3707--3716, 2016.

\bibitem{chabert2006relighting}
Charles-F{\'e}lix Chabert, Per Einarsson, Andrew Jones, Bruce Lamond, Wan-Chun
  Ma, Sebastian Sylwan, Tim Hawkins, and Paul Debevec.
\newblock Relighting human locomotion with flowed reflectance fields.
\newblock In {\em ACM SIGGRAPH 2006 Sketches}, pages 76--es. 2006.

\bibitem{Tran2018}
Du~Tran, Heng Wang, Lorenzo Torresani, Jamie Ray, Yann LeCun, and Manohar
  Paluri.
\newblock {A Closer Look at Spatiotemporal Convolutions for Action
  Recognition}.
\newblock In {\em 2018 IEEE/CVF Conference on Computer Vision and Pattern
  Recognition}, pages 6450--6459. IEEE, jun 2018.

\bibitem{Taniai2018}
Tatsunori Taniai and Takanori Maehara.
\newblock {Neural Inverse Rendering for General Reflectance Photometric
  Stereo}.
\newblock {\em 35th International Conference on Machine Learning, ICML 2018},
  11:7731--7740, feb 2018.

\bibitem{Ikehata2014}
Satoshi Ikehata and Kiyoharu Aizawa.
\newblock {Photometric Stereo Using Constrained Bivariate Regression for
  General Isotropic Surfaces}.
\newblock In {\em 2014 IEEE Conference on Computer Vision and Pattern
  Recognition}, pages 2187--2194. IEEE, jun 2014.

\bibitem{Ikehata2012}
Satoshi Ikehata, David Wipf, Yasuyuki Matsushita, and Kiyoharu Aizawa.
\newblock {Robust photometric stereo using sparse regression}.
\newblock In {\em 2012 IEEE Conference on Computer Vision and Pattern
  Recognition}, number~2, pages 318--325. IEEE, jun 2012.

\bibitem{chen2020deep}
Guanying Chen, Kai Han, Boxin Shi, Yasuyuki Matsushita, and Kwan-Yee~Kenneth
  Wong.
\newblock Deep photometric stereo for non-lambertian surfaces.
\newblock {\em IEEE Transactions on Pattern Analysis and Machine Intelligence},
  2020.

\bibitem{logothetis2020px}
Fotios Logothetis, Ignas Budvytis, Roberto Mecca, and Roberto Cipolla.
\newblock Px-net: Simple, efficient pixel-wise training of photometric stereo
  networks.
\newblock {\em arXiv preprint arXiv:2008.04933}, 2020.

\end{thebibliography}
	
\end{document}